\documentclass{article}
\usepackage{spconf,amsmath,graphicx}
\usepackage{tabularx,booktabs}
\usepackage{multirow}
\usepackage{color}
\usepackage{ragged2e}
\usepackage{placeins}
\usepackage{amssymb}

\usepackage{stackengine}
\def\delequal{\mathrel{\ensurestackMath{\stackon[1pt]{=}{\scriptstyle\Delta}}}}
\title{Is Multi-Task Learning an Upper Bound for Continual Learning?}
\name{Zihao Wu$^1$, Huy Tran$^1$, Hamed Pirsiavash$^2$, and Soheil Kolouri$^1$\thanks{This work was supported by the Defense Advanced Research Projects Agency (DARPA) under Contract No. HR00112190135.}}
\address{{1: Department of Computer Science, Vanderbilt University, Nashville, TN},\\2: Department of Computer Science, University of California, Davis, CA}
\begin{document}
%

\maketitle
\begin{abstract}
Continual and multi-task learning are common machine learning approaches to learning from multiple tasks. The existing works in the literature often assume multi-task learning as a sensible performance upper bound for various continual learning algorithms. While this assumption is empirically verified for different continual learning benchmarks, it is not rigorously justified. Moreover, it is imaginable that when learning from multiple tasks, a small subset of these tasks could behave as adversarial tasks reducing the overall learning performance in a multi-task setting. In contrast, continual learning approaches can avoid the performance drop caused by such adversarial tasks to preserve their performance on the rest of the tasks, leading to better performance than a multi-task learner. This paper proposes a novel continual self-supervised learning setting, where each task corresponds to learning an invariant representation for a specific class of data augmentations. In this setting, we show that continual learning often beats multi-task learning on various benchmark datasets, including MNIST, CIFAR-10, and CIFAR-100. 
\end{abstract}
\begin{keywords}
Continual Learning, Multi-Task Learning, Self Supervised Learning
\end{keywords}
\section{Introduction}
\label{sec:intro}

Modern Machine Learning (ML) is rapidly moving away from single-task experts towards foundational models that can generalize to multiple tasks. Multi-task and continual learning are the two commonly used paradigms in machine learning when learning from a multitude of tasks. Multi-task learning (MTL) assumes simultaneous access to independent and identically distributed (i.i.d) samples from the joint distribution over all tasks and trains the ML model on this joint distribution.
However, in many practical settings, e.g., autonomous driving, one deals with an input data stream and joint training on the ever-growing data and its ever-changing distribution poses a major challenge to MTL. In contrast, continual learning (CL) assumes that the ML model can only have access to one task at a time and learns them sequentially. We note that, appropriately, MTL is sometimes referred to as joint training, while CL is referred to as sequential/incremental training. A dire consequence of not having access to i.i.d. samples from the joint distribution in CL is the catastrophic forgetting phenomenon, which refers to the loss of performance on previous tasks while learning new ones.

The research community has recently proposed a plethora of approaches for overcoming catastrophic forgetting in the continual learning of deep neural networks. One can broadly categorize these methods into: 1) regularization-based approaches that penalize large changes to important parameters of previous tasks \cite{kirkpatrick2017overcoming,zenke2017continual,aljundi2017memory,kolouri2020sliced,li2021lifelong}, and 2) memory replay and rehearsal-based approaches \cite{shin2017continual,rolnick2019experience,rostami2020generative,farajtabar2020orthogonal}, and 3) architectural methods that rely on model expansion, parameter isolation, and masking \cite{schwarz2018progress,mallya2018packnet,mallya2018piggyback,wortsman2020supermasks}. Others have studied brain-inspired mechanisms \cite{kudithipudi2022biological} that allow for continual learning in mammalian brains and how to leverage them for continual machine learning. Interestingly, in nearly all existing approaches, the common baselines are MTL (i.e., joint training) as an upper bound and naive sequential training, leading to catastrophic forgetting, as a lower bound. 


In this paper, we argue that MTL, while being a valuable baseline, is not necessarily an upper bound for CL. We observe that interference in learning is not unique to the CL framework and can happen in MTL. For instance, adversarial examples/tasks, whether optimized or occurring naturally, could significantly reduce the performance of a multi-task learner during training. While CL methods that overcome catastrophic forgetting could avoid the performance drop caused by such adversarial examples/tasks in favor of preserving the performance on non-adversarial examples/tasks. Notably, one can argue that this effect does not happen in non-adversarial settings, as evident in the results on most CL benchmark problems indicating MTL as an upper bound for CL; hence, it could be of limited interest to the community. In response to this criticism, we show that our observation is not unique to adversarial settings and can naturally happen in CL. We introduce the continual learning of augmentation invariant representations as a continual self-supervised learning (SSL) problem. We show that, in this setting, CL often outperforms its MTL counterpart on various benchmark datasets, including MNIST, CIFAR10, and CIFAR100.


\section{Related Work}

\noindent{\bf Self Supervised Learning (SSL)} has become the dominant paradigm in unsupervised learning of visual representations. State-of-the-art SSL algorithms share the underlying theme of learning representations that are invariant to input augmentations (i.e., distortions of input images). These methods can be generally categorized into contrastive and non-contrastive approaches. Contrastive learning algorithms encourage similar embeddings for different augmentations of the same image while distancing embeddings from other images (i.e., the negative samples). Non-contrastive methods, on the other hand, remove the need for explicit negative samples, and only enforce similar embeddings for different augmentations of positive samples. In this paper, we utilize Barlow Twins \cite{zbontar2021barlow}, a non-contrastive SSL algorithm that relies on redundancy reduction, as our core SSL algorithm.

\noindent{\bf Continual Learning (CL)} concerns learning ML models from non-stationary input data streams. The majority of existing research in the literature focuses on overcoming catastrophic forgetting in data, class, or domain incremental learning settings \cite{van2019three}. Notably, the standard benchmarks in the CL literature consider supervised continual learning, while the CL setting is better suited for learning from unlabeled data. Recently, there have been several works that focus on continual self-supervised learning \cite{cha2021co2l,gomez2022continually,fini2022self}, which hold promise for the next generation of CL algorithms. In this paper, we leverage the CaSSLe framework \cite{fini2022self} for continual SSL. In contrast to the existing works in the literature, however, we do not consider incremental learning of classes or domains as our tasks. We observe that the choice of augmentation used for learning invariant representations in SSL changes the input distribution of a CL model. Hence, we treat each type of augmentation as a task and continually train an SSL model on these tasks/augmentations. In other words, we incrementally learn invariant representations to different types of input augmentations. We show that in incremental invariant representation learning, MTL's performance is inferior to CL approaches. 

Lastly, there are emerging works providing connections between MTL and CL methods, e.g., Mirzadeh et al. \cite{mirzadeh2020linear}, which relate the solutions of multitask and continual learning. Such approaches, however, show these connections on standard CL benchmark datasets for which MTL provides an ``upper bound'' for CL. Our work encourages the research community to consider alternative learning settings in which MTL is not an upper bound for CL.

\section{Method}

Let $X=\{x_i\}_{i=1}^n$ denote a batch of images sampled from the dataset, $\mathcal{T}_t$ denote the distribution of image augmentation for the $t$'th task (e.g., cropping), and $f_t$ denote the deep SSL model consisting of a backbone and a projection head. At task $t$,  two distorted views of the input batch are generated as $Y^A=\{y^A_i=T^A_i(x_i)\}_{i=1}^n$ and $Y^B=\{y^B_i=T^B_i(x_i)\}_{i=1}^n$, where $T^A_i,T^B_i\sim \mathcal{T}_t$ for $\forall i$. The distorted batches are then fed to the function $f_t$ to produce batches of d-dimensional embeddings $Z^A, Z^B\in \mathbb{R}^{n\times d}$ (we use $Z^A=f_t(Y^A)$, and $Z^B=f_t(Y^B)$). We adopt the Barlow Twins framework \cite{zbontar2021barlow} and enforce the cross correlation between $Z^A$ and $Z^B$ to be close to the identity matrix by minimizing the following loss,  
\begin{align}
    \mathcal{L}_{BT}(Z^A,Z^B)\delequal \sum_i (1-\mathcal{C}_{ii})^2 +\lambda \sum_i\sum_{j\neq i} \mathcal{C}^2_{ij}
\end{align}
where the first term enforces invariance to augmentation, the second term enforces redundancy reduction, $\lambda$ is a positive regularization coefficient that quantifies the relative importance of each term, and $\mathcal{C}\in \mathbb{R}^{d\times d}$ is the cross correlation matrix along the batch dimension. 

In our continual self-supervised learning setting, the input data is fixed and the tasks are formed by changing the distribution of data augmentations. In short, we use a set of $T$ unique types of augmentations, $\{\mathcal{T}_t\}_{t=1}^T$, e.g., cropping, jittering, Gaussian noise, etc., and incrementally learn invariant representations to these augmentations, only using augmentation $\mathcal{T}_t$ when learning the $t$'th task. 

\begin{figure}[t!]
    \centering
    \includegraphics[width=\columnwidth]{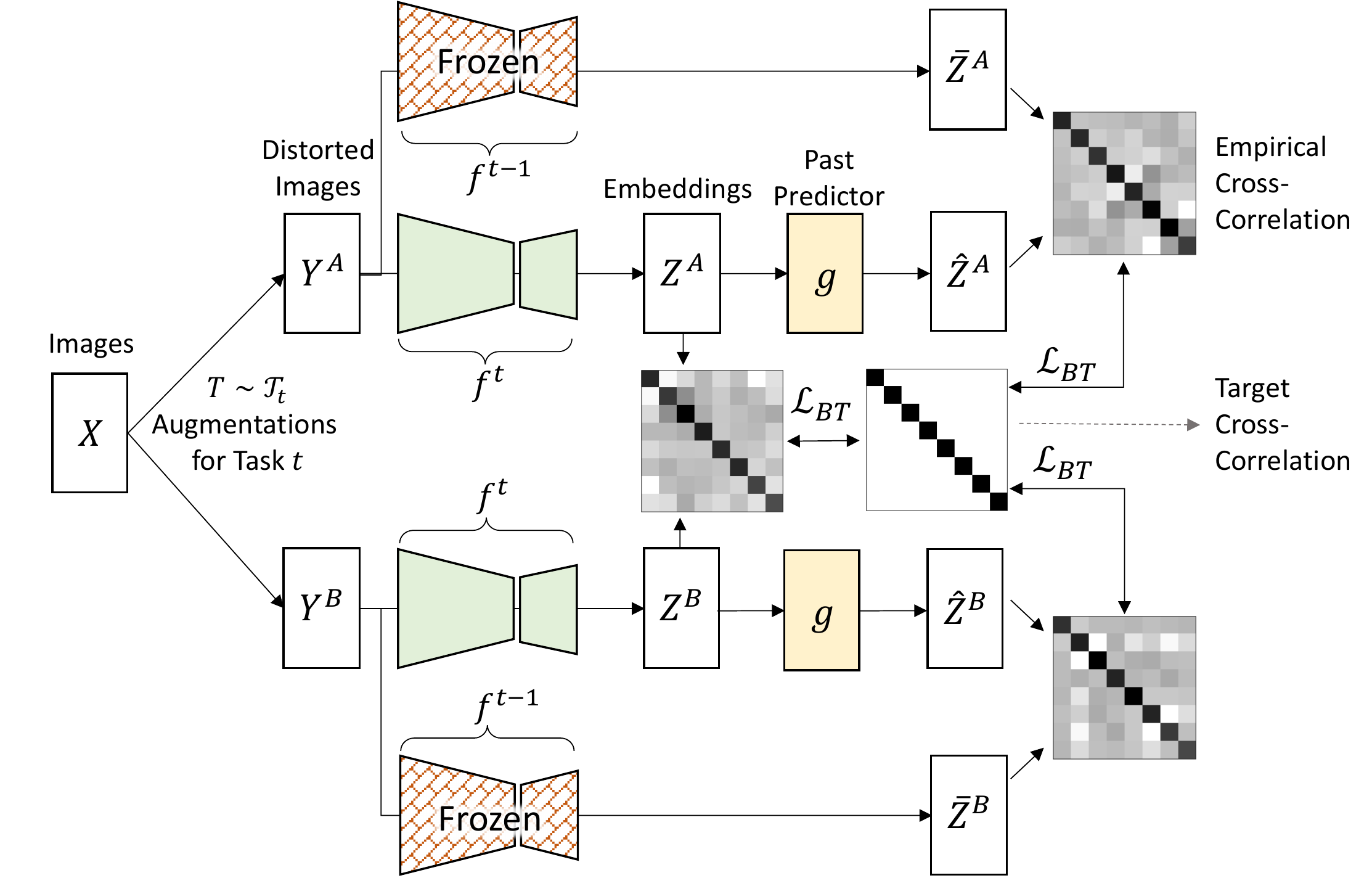}
    \caption{While learning Task $t$, we first generate two views, $Y^A$ and $Y^B$, of the input batch of images, $X$, based on augmentations sampled from $\mathcal{T}_t$, and embed them via the current model, $f_t$, leading to $Z^A=f_t(Y^A)$ and $Z^B=f_t(Y^B)$, and the previous model $f_{t-1}$ (frozen) leading to $\bar{Z}^A=f_{t-1}(Y^A)$ and $\bar{Z}^B=f_{t-1}(Y^B)$. We then predict the embedding of the previous model using the current embedding via a past predictor model, $g$, leading to $\hat{Z}^A=g(Z^A)$ and $\hat{Z}^B=g(Z^B)$. Lastly, we enforce the correlation matrices between: 1) $Z^A$ and $Z^B$, 2) $\bar{Z}^A$ and $\hat{Z}^A$, and 3) $\bar{Z}^B$ and $\hat{Z}^B$ to be close to the identity matrix, simultaneously enforcing invariant representations, redundancy reduction, and information accumulation.}
    \label{fig:architecture}
\end{figure}

To avoid catastrophic forgetting, we use the CaSSLe framework \cite{fini2022self}, which is designed for continual SSL. Similar to distillation-based approaches in CL, CaSSLe carries a frozen copy of the model trained on the previous task, $f_{t-1}$. However, unlike distillation-based approaches that require representations not to change while training the new model, CaSSLe requires the output of the new model to be predictive of the old model (i.e., no information should be lost). Formally, for Task $t$, let $\bar{Z}^A=f_{t-1}(Y^A)$ and $\bar{Z}^B=f_{t-1}(Y^B)$ be the outputs of the old model on different views of the images augmented with $\mathcal{T}_t$, let $g:\mathbb{R}^d\rightarrow \mathbb{R}^d$ denote a past prediction model that is to predict the output of the previous model from the output of the current model, and let $\hat{Z}^A=g(Z^A)$ denote the prediction of past embedding. Then, CaSSLe uses the following loss:
\begin{align}
    \mathcal{L}=\underbrace{\mathcal{L}_{BT}(Z^A,Z^B)}_{SSL}+\gamma(\underbrace{\mathcal{L}_{BT}(\bar{Z}^A,\hat{Z}^A)+\mathcal{L}_{BT}(\bar{Z}^B,\hat{Z}^B)}_{CL})
\end{align}
where the first term is the SSL loss, while the second and third terms are the distillation terms enforcing no information loss during continual training of the SSL model, and $gamma$ is the distillation weight. Figure \ref{fig:architecture} depicts the architecture of CaSSLe used in our work. 


\section{Numerical Experiments}

\subsection{Experimental Setup}

Our goal is to incrementally learn invariant representations for different types of augmentations, in a continual SSL setting. 
We demonstrate our approach on three benchmark datasets, namely CIFAR-10, CIFAR-100, and MNIST. 
For each dataset, we select five types of augmentations to create a CL experiment with five tasks. For each dataset, each augmentation is treated as a task in the task sequence for training our models. For CIFAR-10/100 we select cropping, flipping, jittering, adding Gaussian noises, gray scaling, while for MNIST we select cropping, perspective shifting, affine shifting, rotation, and adding Gaussian noise. Note that order of tasks, i.e., the curriculum, matters for our continual learner. Thus, we generate five random curricula for CIFAR-10/100 and another set of five random curricula for MNIST. These curricula are shown in Table \ref{tab:curricula}. 

\subsection{Training Details}
For our model architecture, $f$ in Figure \ref{fig:architecture}, we utilize ResNet18 as our backbone along with a projector network that has three linear layers, each with 2048 output units. For the past predictor module, $g$, we use two linear layers, each with 2048 output units. All these linear layers except the output layer are followed by a batch normalization layer and rectified linear units. We train the model with a batch size of 256 using the Adam optimizer \cite{kingma2014adam}, distillation weight of $\gamma=0.5$, the Barlow Twins regularization coefficient of $\lambda=0.005$, and a fixed learning rate of $5e^{-4}$ for all three datasets. We train the models for 100 epochs/task for CIFAR-10/100 and 50 epochs/task for MNIST. We randomly split five percent of each training set as validation set.

\subsection{Evaluation Details}

We follow the common practice in SSL and evaluate the trained models via a linear head. In short, for each dataset, we train a linear classifier on top of the fixed representations of our self-supervised ResNet18, and report the classification accuracy. We perform linear evaluation at the end of learning each task. As the baseline, we report the MTL performance on the same number of tasks (jointly trained) trained over an equivalent number of epochs to the continual learner (i.e, for a sequence of five tasks trained on MNIST, MTL is trained on 250 epochs). We repeat each experiment five times and report the average performance throughout our Results section. 


\begin{table}[t!]
\caption{Training curricula, where GN refers to Gaussian noise, Rot. to rotation, and Pers. to perspective shifting.}
\begin{tabular}{@{\extracolsep{4pt}}ccccccc@{}}
\toprule[1.5pt]
&  &
\multicolumn{5}{c}{Curriculum}\\
\cline{3-3}
\cline{4-4}
\cline{5-5}
\cline{6-6}
\cline{7-7}
& Task & 1 & 2 & 3 & 4 & 5 \\
\hline
\multirow{5}{*}{\rotatebox{90}{CIFARs}} &
A1   &Crop    &Crop    &Jitter    &Jitter     &Gray\\
&A2  &Flip    &Jitter  &Flip      &GN         &GN\\
&A3  &Jitter  &Gray    &GN        &Flip       &Jitter\\
&A4  &GN      &Flip    &Crop      &Gray       &Flip\\
&A5  & Gray   &GN      &Gray      &Crop       &Crop\\
\midrule
\multirow{5}{*}{\rotatebox{90}{MNIST}} &
B1   &Crop    &GN      &Affine    &Affine     &Persp.\\
&B2  &Persp.    &Rot.     &Persp.      &Rot.        &Rot.\\
&B3  &Affine  &Affine  &Rot.       &Persp.       &Affine\\
&B4  &Rot.     &Persp.    &Crop      &GN         &GN\\
&B5  &GN      &Crop    &GN        &Crop       &Crop\\
\bottomrule[1.5pt]
\end{tabular}
\label{tab:curricula}
\end{table}

\begin{table}[t!]
\vspace{-.2in}
\centering
\caption{Training results for individual augmentations.}
\begin{tabular}{@{\extracolsep{4pt}}cccc@{}}
\toprule[1.5pt]
 & 
\multicolumn{3}{c}{Dataset}\\
\cline{2-2}
\cline{3-3}
\cline{4-4}
Augmentation
& CIFAR-10 & CIFAR-100 & MNIST \\
\hline
Flip            &43.74          &20.95          &-\\
Jitter          &42.89          &17.75          &-\\
Gray            &32.49          &9.61          &-\\
Gaussian Noise  &37.96          &15.73          &93.67\\
Crop            &53.60          &28.59           &98.85\\
Perspective     &-              &-              &98.88\\
Affine          &-              &-              &96.85\\
Rotation        &-              &-              &96.26\\
\bottomrule[1.5pt]
\end{tabular}
\label{tab:single}
\end{table}

\begin{table*}[t!]
\begin{tabularx}{\textwidth}{@{\extracolsep{4pt}} l l *{10}{c} c @{}}
\toprule[1.5pt]
\multicolumn{1}{l}{} &
\multicolumn{2}{l}{} &
\multicolumn{2}{c}{Curriculum 1} & 
\multicolumn{2}{c}{Curriculum 2} &
\multicolumn{2}{c}{Curriculum 3} & 
\multicolumn{2}{c}{Curriculum 4} & 
\multicolumn{2}{c}{Curriculum 5} \\
\cline{4-5}
\cline{6-7}
\cline{8-9}
\cline{10-11}
\cline{12-13}
&
Task Sequence
&
&
MTL & CL &
MTL & CL &
MTL & CL &
MTL & CL &
MTL & CL \\
\hline
\multirow{5}{*}{\rotatebox{90}{CIFAR-10}}
    &$A_1$               &&53.60           &53.60          &53.60          &53.60                        &42.89   &42.89   &42.89    &42.89     &32.49    &32.49   \\ 
    &$A_1,A_2$            &&50.52    &\textbf{52.62}    &53.76    &\textbf{60.83}    &\textbf{45.60}    &44.15   &41.36    &\textbf{42.76}     &\textbf{34.64}   &34.28   \\ 
    &$A_1,A_2,A_3$         &&53.96    &\textbf{58.93}      &51.7  &\textbf{60.84}    &44.38   &\textbf{44.72}  &\textbf{44.38}   &43.85    &\textbf{38.44}    &38.18    \\ 
    &$A_1,A_2,A_3,A_4$      &&\textbf{59.57}    &57.83      &50.45   &\textbf{60.67}   &\textbf{59.57}   &53.32 &40.17   &\textbf{44.00}    &40.17    &\textbf{45.32}    \\ 
    &$A_1,A_2,A_3,A_4,A_5$   &&50.55    &\textbf{57.38}      &50.55   &\textbf{60.90}   &50.55  &\textbf{52.24}  &50.55   &\textbf{52.84}    &50.55    &\textbf{57.33}    \\  
\midrule
\multirow{5}{*}{\rotatebox{90}{CIFAR-100}} 
    &$A_1$               &&28.59            &28.59              &28.59    &28.59            &17.75   &17.75   &17.75    &17.75     &9.61    &9.61    \\ 
    &$A_1,A_2$            &&26.54            &\textbf{27.27}     &27.09    &\textbf{34.88}   &\textbf{19.25}    &18.30   &16.29    &\textbf{17.78}     &\textbf{11.57}   &10.05   \\ 
    &$A_1,A_2,A_3$         &&27.74            &\textbf{33.59}     &23.79     &\textbf{34.09}   &\textbf{19.10}   &18.78  &\textbf{19.10}   &18.66    &\textbf{13.89}    &11.81    \\ 
    &$A_1,A_2,A_3,A_4$      &&\textbf{33.98}   &32.22              &23.30    &\textbf{33.47}   &\textbf{33.98}   &27.91 &14.94   &\textbf{18.31}    &\textbf{14.94}    &14.72    \\ 
    &$A_1,A_2,A_3,A_4,A_5$   &&25.76            &\textbf{30.90}     &25.76    &\textbf{33.77}  &25.76  &\textbf{26.47}  &25.76   &\textbf{27.70}    &25.76    &\textbf{29.20}    \\  
\midrule[1.5pt]

\multirow{5}{*}{\rotatebox{90}{MNIST}} 
    &$B_1$              &&98.85       &98.85   &93.67    &93.67  &96.85   &96.85   &96.85    &96.85     &98.88    &98.88    \\ 
    &$B_1,B_2$            &&98.92    &\textbf{99.23}    &95.80    &\textbf{96.82}   &\textbf{99.06}    &98.92   &96.75    &\textbf{96.84}     &\textbf{99.02}   &99.01   \\ 
    &$B_1,B_2,B_3$         &&99.19    &\textbf{99.26}      &96.46  &\textbf{97.64}  &98.31   &\textbf{98.9}  &98.32   &\textbf{98.81}    &98.32    &\textbf{99.06}    \\ 
    &$B_1,B_2,B_3,B_4$      &&99.13    &\textbf{99.28}      &98.98   &\textbf{99.05} &\textbf{99.13}   &\textbf{99.13} &\textbf{98.98}   &98.78    &\textbf{98.98}    &98.90    \\ 
    &$B_1,B_2,B_3,B_4,B_5$   &&99.14    &\textbf{99.15}      &99.14   &\textbf{99.25}  &\textbf{99.14}  &99.02  &99.14   &\textbf{99.26}    &99.14    &\textbf{99.24}    \\  
\bottomrule[1.5pt]
\end{tabularx}
\caption{Comparison between our CL method and the MTL baseline across curricula. Tasks for each sequence are selected based on curricula as defined in Table 1.}
\label{tab:full}
\end{table*}

\subsection{Results}

We start by quantifying the contribution of each individual augmentation to SSL on the three datasets. We train our model using the Barlow Twins loss using only one type of augmentation, and report the linear evaluations averaged over five runs in Table \ref{tab:single}. In accordance with prior published work in the literature we see that flip and crop augmentations contribute the most to learning CIFAR-10/100, while for MNIST perspective shifting and crop have the largest contribution. 

Next, we follow the random curricula in Table \ref{tab:curricula} and train the model sequentially on different augmentations. At the end of each task, we perform linear evaluation. For comparison, we perform MTL and jointly train the model on the augmentations. We repeat each experiment five times and report the average performance for all datasets, for all curricula, and for both CL and MTL in Table \ref{tab:full}. Note that, CL and MTL are identical when learning from a single task.  Our results indicate that MTL is no longer an upper bound for CL, and in fact, CL more consistently results in better or comparable performance to that of MTL. 

We point out that learning invariant representations introduces naturally adversarial tasks. For instance, for CIFAR-10, SSL with $A_1$ for Curriculum 1 (Crop) results in a $\%53.60$ linear evaluation performance, while joint training with $A_1$ and $A_2$ (Flip) results in $\%50.52$, which means that there is negative transfer between the two tasks (i.e., the tasks are naturally adversarial to one another). Interestingly, CL also suffers from a performance drop but the effect is less severe, from $\%53.60$ to $\%52.62$. This phenomenon happens many times throughout our experiment, as it is evident in Table \ref{tab:full}. 

\begin{table}[t!]
\vspace{-.1in}
\centering
\caption{Average negative transfer.}
\begin{tabular}{@{\extracolsep{4pt}}cccc@{}}
\toprule[1.5pt]
 & 
\multicolumn{3}{c}{Dataset}\\
\cline{2-2}
\cline{3-3}
\cline{4-4}
Neg. Transfer
& CIFAR-10 & CIFAR-100 & MNIST \\
\hline
MTL         & -3.14          &-2.78          &-0.32\\
CL          &\bf{-0.44}         & \bf{-0.76}   &\bf{-0.07} \\
\bottomrule[1.5pt]
\end{tabular}
\label{tab:negTrans}
\end{table}

Lastly, we calculate the average negative transfer for MTL and CL for all datasets and over all curricula. Briefly, we calculate the number of times when there is a performance drop from  $A_{1:i}$ to $A_{1:i+1}$ and report the average performance drop for MTL and CL. Table \ref{tab:negTrans} shows the calculated average negative transfer. As can be seen, MTL suffers from higher performance drop compared to CL on all datasets.

\section{Conclusion}
\label{sec:conclusion}
In this paper, we sought to answer the question: does multi-task learning (MTL) provide an upper bound for continual learning (CL)? We posed the problem of learning invariant representations in self-supervised learning (SSL) as a CL problem, where the learner must incrementally learn invariant representations to different types of augmentations. We showed that such a continual learning setting poses naturally adversarial (i.e., conflicting) tasks that lead to a performance drop for MTL. We then showed that CL could avoid this performance drop and provide superior performance compared to MTL on continual self-supervised learning on various benchmark datasets, namely CIFAR-10, CIFAR-100, and MNIST. In the future, we will leverage the observations in this paper to design better self-supervised learning algorithms on large-scale image datasets. 

\bibliographystyle{IEEEbib}
\bibliography{CLvsMTL}

\begin{thebibliography}{10}

\bibitem{kirkpatrick2017overcoming}
James Kirkpatrick, Razvan Pascanu, Neil Rabinowitz, Joel Veness, Guillaume
  Desjardins, Andrei~A Rusu, Kieran Milan, John Quan, Tiago Ramalho, Agnieszka
  Grabska-Barwinska, et~al.,
\newblock ``Overcoming catastrophic forgetting in neural networks,''
\newblock {\em Proceedings of the national academy of sciences}, vol. 114, no.
  13, pp. 3521--3526, 2017.

\bibitem{zenke2017continual}
Friedemann Zenke, Ben Poole, and Surya Ganguli,
\newblock ``Continual learning through synaptic intelligence,''
\newblock in {\em International Conference on Machine Learning}. PMLR, 2017,
  pp. 3987--3995.

\bibitem{aljundi2017memory}
Rahaf Aljundi, Francesca Babiloni, Mohamed Elhoseiny, Marcus Rohrbach, and
  Tinne Tuytelaars,
\newblock ``Memory aware synapses: Learning what (not) to forget,''
\newblock in {\em ECCV}, 2018.

\bibitem{kolouri2020sliced}
Soheil Kolouri, Nicholas~A Ketz, Andrea Soltoggio, and Praveen~K Pilly,
\newblock ``Sliced cramer synaptic consolidation for preserving deeply learned
  representations,''
\newblock in {\em International Conference on Learning Representations}, 2020.

\bibitem{li2021lifelong}
Haoran Li, Aditya Krishnan, Jingfeng Wu, Soheil Kolouri, Praveen~K Pilly, and
  Vladimir Braverman,
\newblock ``Lifelong learning with sketched structural regularization,''
\newblock 2021, vol. 157 of {\em Proceedings of Machine Learning Research},
  {PMLR}.

\bibitem{shin2017continual}
Hanul Shin, Jung~Kwon Lee, Jaehong Kim, and Jiwon Kim,
\newblock ``Continual learning with deep generative replay,''
\newblock in {\em Proceedings of the 31st International Conference on Neural
  Information Processing Systems}, 2017, pp. 2994--3003.

\bibitem{rolnick2019experience}
David Rolnick, Arun Ahuja, Jonathan Schwarz, Timothy Lillicrap, and Gregory
  Wayne,
\newblock ``Experience replay for continual learning,''
\newblock {\em Advances in Neural Information Processing Systems}, vol. 32,
  2019.

\bibitem{rostami2020generative}
Mohammad Rostami, Soheil Kolouri, Praveen Pilly, and James McClelland,
\newblock ``Generative continual concept learning,''
\newblock in {\em Proceedings of the AAAI Conference on Artificial
  Intelligence}, 2020, vol.~34, pp. 5545--5552.

\bibitem{farajtabar2020orthogonal}
Mehrdad Farajtabar, Navid Azizan, Alex Mott, and Ang Li,
\newblock ``Orthogonal gradient descent for continual learning,''
\newblock in {\em International Conference on Artificial Intelligence and
  Statistics}. PMLR, 2020, pp. 3762--3773.

\bibitem{schwarz2018progress}
Jonathan Schwarz, Wojciech Czarnecki, Jelena Luketina, Agnieszka
  Grabska-Barwinska, Yee~Whye Teh, Razvan Pascanu, and Raia Hadsell,
\newblock ``Progress \& compress: A scalable framework for continual
  learning,''
\newblock in {\em International Conference on Machine Learning}. PMLR, 2018,
  pp. 4528--4537.

\bibitem{mallya2018packnet}
Arun Mallya and Svetlana Lazebnik,
\newblock ``Packnet: Adding multiple tasks to a single network by iterative
  pruning,''
\newblock in {\em Proceedings of the IEEE conference on Computer Vision and
  Pattern Recognition}, 2018, pp. 7765--7773.

\bibitem{mallya2018piggyback}
Arun Mallya, Dillon Davis, and Svetlana Lazebnik,
\newblock ``Piggyback: Adapting a single network to multiple tasks by learning
  to mask weights,''
\newblock in {\em Proceedings of the European Conference on Computer Vision
  (ECCV)}, 2018, pp. 67--82.

\bibitem{wortsman2020supermasks}
Mitchell Wortsman, Vivek Ramanujan, Rosanne Liu, Aniruddha Kembhavi, Mohammad
  Rastegari, Jason Yosinski, and Ali Farhadi,
\newblock ``Supermasks in superposition,''
\newblock {\em Advances in Neural Information Processing Systems}, vol. 33, pp.
  15173--15184, 2020.

\bibitem{kudithipudi2022biological}
Dhireesha Kudithipudi, Mario Aguilar-Simon, Jonathan Babb, Maxim Bazhenov,
  Douglas Blackiston, Josh Bongard, Andrew~P Brna, Suraj Chakravarthi~Raja,
  Nick Cheney, Jeff Clune, et~al.,
\newblock ``Biological underpinnings for lifelong learning machines,''
\newblock {\em Nature Machine Intelligence}, vol. 4, no. 3, pp. 196--210, 2022.

\bibitem{zbontar2021barlow}
Jure Zbontar, Li~Jing, Ishan Misra, Yann LeCun, and St{\'e}phane Deny,
\newblock ``Barlow twins: Self-supervised learning via redundancy reduction,''
\newblock in {\em International Conference on Machine Learning}. PMLR, 2021,
  pp. 12310--12320.

\bibitem{van2019three}
Gido~M Van~de Ven and Andreas~S Tolias,
\newblock ``Three scenarios for continual learning,''
\newblock {\em arXiv preprint arXiv:1904.07734}, 2019.

\bibitem{cha2021co2l}
Hyuntak Cha, Jaeho Lee, and Jinwoo Shin,
\newblock ``Co2l: Contrastive continual learning,''
\newblock in {\em Proceedings of the IEEE/CVF International Conference on
  Computer Vision}, 2021, pp. 9516--9525.

\bibitem{gomez2022continually}
Alex Gomez-Villa, Bartlomiej Twardowski, Lu~Yu, Andrew~D Bagdanov, and Joost
  van~de Weijer,
\newblock ``Continually learning self-supervised representations with projected
  functional regularization,''
\newblock in {\em Proceedings of the IEEE/CVF Conference on Computer Vision and
  Pattern Recognition}, 2022, pp. 3867--3877.

\bibitem{fini2022self}
Enrico Fini, Victor G~Turrisi da~Costa, Xavier Alameda-Pineda, Elisa Ricci,
  Karteek Alahari, and Julien Mairal,
\newblock ``Self-supervised models are continual learners,''
\newblock in {\em Proceedings of the IEEE/CVF Conference on Computer Vision and
  Pattern Recognition}, 2022, pp. 9621--9630.

\bibitem{mirzadeh2020linear}
Seyed~Iman Mirzadeh, Mehrdad Farajtabar, Dilan Gorur, Razvan Pascanu, and
  Hassan Ghasemzadeh,
\newblock ``Linear mode connectivity in multitask and continual learning,''
\newblock in {\em International Conference on Learning Representations}, 2020.

\bibitem{kingma2014adam}
Diederik~P Kingma and Jimmy Ba,
\newblock ``Adam: A method for stochastic optimization,''
\newblock {\em arXiv preprint arXiv:1412.6980}, 2014.

\end{thebibliography}

\end{document}